%% file: Template.tex
\documentclass{article}
\usepackage{spconf,amsmath,graphicx}
\DeclareMathOperator*{\argmax}{arg\,max}

\usepackage{times}
\usepackage{latexsym}

\usepackage{graphicx}
\usepackage{amsfonts}

\usepackage[table]{xcolor}
\usepackage{array}
\usepackage{url}

\usepackage{algorithm}
\usepackage{algorithmicx,algpseudocode}

\usepackage{enumitem}
\setlist{nosep}

\usepackage{multirow}
\usepackage{tabularx}
\usepackage{multicol}
\usepackage{multirow}
\usepackage{diagbox}
\usepackage{amsmath}
\usepackage{mathtools}
\usepackage{bm}
\usepackage{booktabs}
\usepackage{float}
\usepackage{graphicx}
\usepackage{caption}
\usepackage[position=top]{subfig}
\usepackage[font=footnotesize,labelfont=bf]{caption}
\usepackage{enumitem}
\usepackage{makecell}
\newcolumntype{x}[1]{>{\centering\arraybackslash}p{#1}}

\usepackage{multirow,bigdelim,dcolumn,booktabs}

\usepackage{arydshln}

\usepackage{tikz}
\usetikzlibrary{matrix,decorations.pathreplacing,calc}

\pgfkeys{tikz/mymatrixenv/.style={decoration=brace,every left delimiter/.style={xshift=3pt},every right delimiter/.style={xshift=-3pt}}}
\pgfkeys{tikz/mymatrix/.style={matrix of math nodes,left delimiter=[,right delimiter={]},inner sep=2pt,column sep=1em,row sep=0.5em,nodes={inner sep=0pt}}}
\pgfkeys{tikz/mymatrixbrace/.style={decorate,thick}}

\title{Optimizing Neural Network Hyperparameters \\with Gaussian Processes for Dialog Act Classification}

\name{Franck Dernoncourt*\thanks{*These authors contributed equally to this work.}, Ji Young Lee*\footnotemark[1]}
\address{MIT CSAIL\\ Cambridge, MA, USA\\\{\tt francky,jjylee\}@mit.edu}

\begin{document}
\maketitle
\begin{tikzpicture}[remember picture, overlay]
\node at ($(current page.north) + (-0in,-0.5in)$) {Accepted as a conference paper at IEEE SLT 2016};
\end{tikzpicture}
\vspace{-0.70cm}
\begin{abstract}
Systems based on artificial neural networks (ANNs) have achieved state-of-the-art results in many natural language processing tasks. Although ANNs do not require manually engineered features, ANNs have many hyperparameters to be optimized. The choice of hyperparameters significantly impacts models' performances. However, the ANN hyperparameters are typically chosen by manual, grid, or random search, which either requires expert experiences or is computationally expensive. Recent approaches based on Bayesian optimization using Gaussian processes (GPs) is a more systematic way to automatically pinpoint optimal or near-optimal machine learning hyperparameters. Using a previously published ANN model yielding state-of-the-art results for dialog act classification, we demonstrate that optimizing hyperparameters using GP further improves the results, and reduces the computational time by a factor of 4 compared to a random search. Therefore it is a useful technique for tuning ANN models to yield the best performances for natural language processing tasks.

\end{abstract}

\input{introduction.tex}

\input{model.tex}

\input{experiments.tex}

\input{results.tex}

\input{conclusion.tex}

\bibliographystyle{IEEEbib}
\newpage
\bibliography{Template}

\end{document}

%% file: introduction.tex
\vspace{0.5cm}
\section{Introduction and related work}

Artificial neural networks (ANNs) have recently shown state-of-the-art results on various NLP tasks including 
language modeling~\cite{mikolov2010recurrent},  named entity recognition~\cite{collobert2011natural,lample2016neural,labeau-loser-allauzen:2015:EMNLP}, text classification~\cite{socher2013recursive,kim2014convolutional,blunsom2014convolutional,lee2016sequential}, question answering~\cite{weston2015towards,wang-nyberg:2015:ACL-IJCNLP}, and machine translation~\cite{bahdanau2014neural,tamura2014recurrent}.
Unlike other popular non-ANN-based machine learning algorithms such as support vector machines (SVMs) and conditional random fields (CRFs), ANNs can automatically learn features that are useful for NLP tasks, thereby requiring no manually engineering features. 

However, ANNs have hyperparameters that need to be tuned in order to achieve the best results. The hyperparameters of ANNs may define either its learning process (e.g., learning rate or mini-batch size) or its architecture (e.g., number of hidden units or layers). ANNs commonly contain over ten hyperparameters~\cite{bengio2012practical}, which makes it challenging to optimize. Therefore, most published ANN-based work on NLP tasks, rely on basic heuristics such as manual or random search, and sometimes do not even optimize hyperparameters. 

Although most of them report state-of-the-art results without optimizing hyperparameters extensively, we argue that the results can be further improved by properly optimizing the hyperparameters. Despite this, one of the main reasons why most previous NLP works do not thoroughly optimize hyperparameters is that it may represent a significant time investment. However, if we optimize them ``efficiently'', we can find hyperparameters that perform well within a reasonable amount of time as shown in this paper. 

Like ANNs, other machine learning algorithms also have hyperparameters. The two most widely used methods for hyperparameter optimization of machine learning algorithms are manual or grid search~\cite{bergstra2012random}. Bergstra and Yoshua~\cite{bergstra2012random} show that random search is as good or better than grid search at finding hyperparameters within a small fraction of computation time and suggest that random search is a natural baseline for judging the performance of automatic approaches for tuning the hyperparameters of a learning algorithm.  
However, all above-mentioned methods for tuning hyperparameters have some downsides. Manual search requires human experts or use arbitrary rules of thumb, while grid and random searches are computationally expensive~\cite{NIPS2012_4522}. 

Recently, a more systematic approach based on Bayesian optimization with Gaussian process~(GP) \cite{williams2006gaussian} has been shown to be effective in automatically tuning the hyperparameters of machine learning algorithms, such as latent dirichlet allocation, SVMs, convolutional neural networks~\cite{NIPS2012_4522}, and deep belief networks~\cite{NIPS2011_4443}, as well as tuning the hyperparameters that features may have~\cite{dernoncourt2015gaussian,dernoncourt2016hyperparameter}. In this approach, the model's performance for each hyperparameter combination is modeled as a sample from a GP, resulting in a tractable posterior distribution given previous experiments. Therefore, this posterior distribution is used to find the optimal hyperparameter combination to try next based on the observation. 

\begin{figure*}[!ht]
  \centering
      \includegraphics[width=\textwidth]{{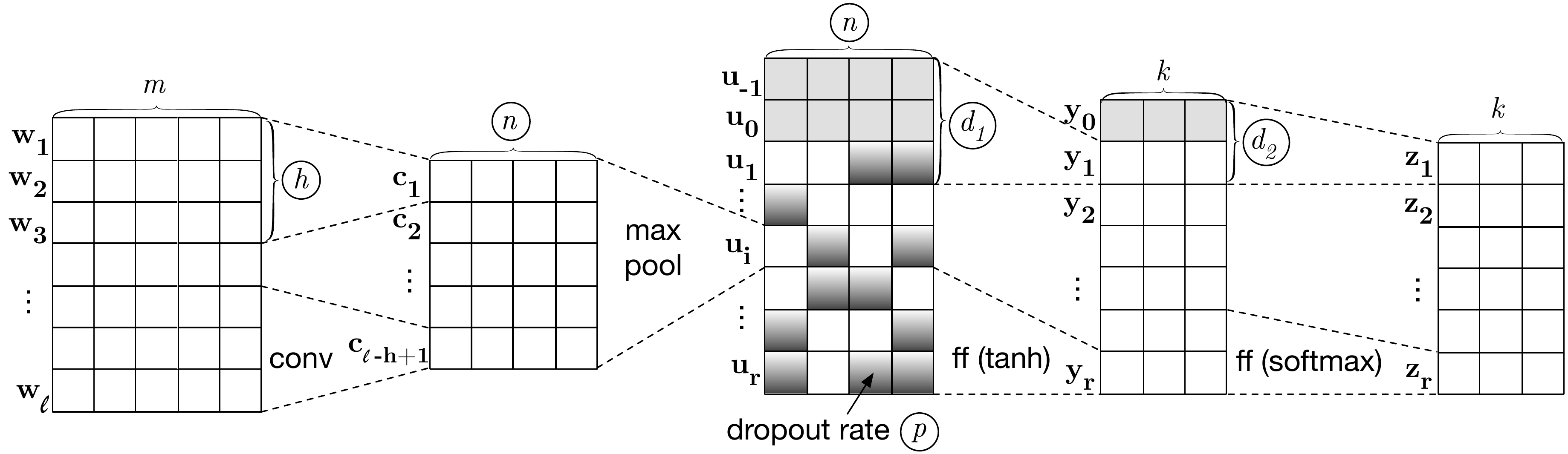}}
  \caption{The ANN model. A sequence of words $\mathbf{w}_{1:\ell}$ corresponding to the $i^{th}$ utterance is transformed into a vector $\mathbf{u}_{i}$ using a CNN, consisting of a convolution layer (conv) and a max pooling layer (max pool). Each utterance is then classified by a two-layer feedforward (ff) network with tanh and softmax activation functions. The hyperparmeters that we optimize are circled: filter size $h$, number of filters $n$, dropout rate $p$, history sizes $d_1, d_2$. In the figure, $h=3,\, n=4,\, p=0.5,\, d_1=3,\, d_2=2.$ The grey rows ($\mathbf{u}_{-1}, \mathbf{u}_0, \mathbf{y}_0$) represent zero paddings. \vspace{-0.0cm}} 

  \label{fig:ann}
\end{figure*}

In this work, we demonstrate the application of Gaussian Process (GP) to optimize ANN hyperparameters on an NLP task, namely dialog act classification~\cite{stolcke2000dialogue}, whose goal is to assign a dialog act to each utterance.
The ANN model in \cite{lee2016sequential} makes a good candidate for hyperparameter optimization since it is a simple model with a few architectural hyperparameters, and the optimized architectural hyperparameters are interpetable and give some insights for the task at hand.  
Using this model, we show that optimizing hyperparameters further improves the state-of-the-art results on two datasets, and reduces the computational time by a factor of 4 compared to a random search.

%% file: model.tex
\section{Methods}
The ANN model for dialog act classification is introduced in \cite{lee2016sequential} and is briefly described in Section~\ref{sec:ann}. The GP used to optimize the hyperparameters of the ANN model is presented in Section~\ref{sec:gp}.
The colon notation $\mathbf{v}_{i:j}$ represents the sequence of vectors $\mathbf{v}_i, \mathbf{v}_{i+1}, \dotsc, \mathbf{v}_j$.

\subsection{ANN model} \label{sec:ann}

\noindent
Each utterance of a dialog is mapped to a vector representation via a CNN (Section~\ref{sec:cnn}).
Each utterance is then sequentially classified by leveraging preceding utterances
(Section~\ref{sec:sequential}). Figure~\ref{fig:ann} gives an overview of the ANN model. 

\subsubsection{Utterance representation via CNN} \label{sec:cnn}
\noindent An utterance of length $\ell$ is represented as the sequence of word vectors $\mathbf{w}_{1:\ell} \in \mathbb{R}^m.$ Given the word vectors, the CNN model produces the \textit{utterance representation} $\mathbf{u} \in \mathbb{R}^n.$

Let $h$ be the size of a \emph{filter}, and the sequence of vectors $\mathbf{v}_{1:h} \in \mathbb{R}^m$ be the corresponding filter matrix. 
A convolution operation on $h$ consecutive word vectors starting from the $t^{th}$ word outputs the scalar \emph{feature} 
$c_t = \textrm{tanh}\left( \sum_{i = 1}^{h} \mathbf{v}_i^T \mathbf{w}_{t+i-1} + b_f \right),$
where $b_f \in \mathbb{R}$ is a bias term.

We perform convolution operations with $n$ different filters, and denote the resulting features as $\mathbf{c}_t \in \mathbb{R}^n,$ each of whose dimensions comes from a distinct filter. 
Repeating the convolution operations for each window of $h$ consecutive words in the utterance, we obtain $\mathbf{c}_{1:\ell-h+1}.$
The utterance representation $\mathbf{u} \in \mathbb{R}^n$ is computed in the max pooling layer, as the element-wise maximum of $\mathbf{c}_{1:\ell-h+1}.$ 
During training, dropout with probability $p$ is applied on this utterance representation $\mathbf{u}.$

The filter size $h,$ the number of filters $n,$ and a dropout probability $p$ are the hyperparameters of this section that we optimize using the GP (Section \ref{sec:gp}).
\subsubsection{Sequential utterance classification} \label{sec:sequential}

\noindent Let $\mathbf{u}_i \in \mathbb{R}^n$ be the utterance representation given by the CNN architecture for the $i^{th}$ utterance in the sequence of length $r$.
The sequence $\mathbf{u}_{1 \, : \, r}$ is input to a two-layer feedforward neural network that classifies each utterance. 
The hyperparameters $d_1, d_2,$ the history sizes used in the first and second layers respectively, are optimized using the GP (Section~\ref{sec:gp}).

The first layer
takes as input $\mathbf{u}_{i-d_1+1 \, : \, i}$ and outputs $\mathbf{y}_i \in \mathbb{R}^k,$ where $k$ is the number of classes for the classification task, i.e. the number of dialog acts.
It uses a tanh activation function. 
Similarly, the second layer takes as input
$\mathbf{y}_{i-d_2+1\,:\,i}$  
 and outputs $\mathbf{z}_i \in \mathbb{R}^k$ with a softmax activation function.

The final output $\mathbf{z}_i$ represents the probability distribution over the set of $k$ classes for the $i^{th}$ utterance: the $j^{th}$ element of $\mathbf{z}_i$ corresponds to the probability that the $i^{th}$ utterance belongs to the $j^{th}$ class. Each utterance is assigned to the class with the highest probability.

\subsection{Hyperparameter optimization using GP} \label{sec:gp}

Let $\mathcal{X}$ be the set of all hyperparameter combinations considered, and let $f: \mathcal{X} \rightarrow \mathbb{R}$ be the function mapping from hyperparameter combinations to a real-valued performance metric (such as F1-score on test set) of a learning algorithm using the given hyperparameter combination. Our interest lies in \emph{efficiently} finding a hyperparameter combination $\mathbf{x} \in \mathcal{X}$ that yields a \emph{near-optimal} performance $f(\mathbf{x})$. In this paper, we use Bayesian optimization of hyperparameters using GP, which we call \emph{GP search}.

\subsubsection{Comparison with other methods}

A grid search is brute-forcefully evaluating $f(\mathbf{x})$ for each ${\mathbf{x} \in \mathcal{X}}$ defined on a grid and then selecting the best one. In a random search, one randomly selects an $\mathbf{x} \in \mathcal{X}$ and evaluates the performance $f(\mathbf{x})$; this process is repeated until an $\mathbf{x}$ with a satisfactory $f(\mathbf{x})$ is found. In a manual search, an expert tries out some hyperparameter combinations based on prior experience until settling on a good one.

In contrast with the other methods mentioned above, a GP search chooses the hyperparameter combination to evaluate next by exploiting all previous evaluations.
To achieve this, we assume the prior distribution on the function $f$ to be a Gaussian process, which allows us to construct a probabilistic model for $f$ using all previous evaluations, by calculating the posterior distribution in a tractable manner. Once the model for $f$ is computed, it is used to choose an optimal hyperparameter combination to evaluate next.

\subsubsection{GP search}

In a GP search, we use a GP to describe a distribution over functions. A GP is defined as a collection of random variables, any finite number of which have a joint Gaussian distribution. A GP $f(\mathbf{x})$ is completely specified by its mean function $m(\mathbf{x})$
and covariance function $k(\mathbf{x}, \mathbf{x}'),$ also called \emph{kernel}, defined as:
\begin{align*}
m(\mathbf{x}) &= \mathbb{E}[f(\mathbf{x})], \\
k(\mathbf{x}, \mathbf{x}') &= \mathbb{E}[(f(\mathbf{x})-m(\mathbf{x}))(f(\mathbf{x}')-m(\mathbf{x}''))].
\end{align*}

In our case $f(\mathbf{x})$ is the F1-score on the test set evaluated for the ANN model using the given hyperparameter combination $\mathbf{x} \in \mathcal{X},$ which is a 5-dimensional vector consisting of filter size $h$, number of filters $n$, dropout rate $p$, and history sizes $d_1, d_2$.

Let ${X = (\mathbf{x}_1,\dotsc,\mathbf{x}_q)},\, {\mathbf{f} = (f(\mathbf{x}_1)\dotsc,f(\mathbf{x}_q))}$ and $X^* = (\mathbf{x}_{q+1},\dotsc,\mathbf{x}_{s}),\, \mathbf{f}^* = (f(\mathbf{x}_{q+1})\dotsc,f(\mathbf{x}_{s}))$ be the training inputs and outputs, and test inputs and outputs, respectively. $X \cup X^* = \mathcal{X},$
 and $X \cap X^* = \emptyset$.   Note that $\mathbf{f}$ is known, and $\mathbf{f}^*$ is unknown. 
The goal is to find the distribution of $\mathbf{f}^*$ given $X^*, X$ and $\mathbf{f},$ in order to select among $X^*$ the hyperparameter combination that is the most likely to yield the highest F1-score.

The joint distribution of $\mathbf{f}$ and $\mathbf{f}^*$ according to the prior is
\vspace{-0.1cm}
\begin{equation*}
	\begin{bmatrix*}[l]
		\mathbf{f} \\ \mathbf{f}^*
	\end{bmatrix*}
	\sim 
	\mathcal{N} \left(  \begin{bmatrix*}[l] 
						\mathbf{m} \\ \mathbf{m}^*
						\end{bmatrix*},
						~
						\begin{bmatrix*}[l]
						K(X,X) & K(X,X^*) \\
						K(X^*,X) & K(X^*,X^*) \\
						\end{bmatrix*}
				\right)
\vspace{-0.1cm}
\end{equation*}
where $\mathbf{m}\,,\,\mathbf{m}^*$ is a vector of the means evaluated at all training and test points respectively, and $K(X,X^*)$ denotes the $q \times q^*$ matrix of the covariances evaluated at all pairs of training and test points, and similarly for $K(X,X),\, K(X^*,X)$ and $K(X^*,X^*)$. 

\vspace{0.1cm}
Conditioning the joint Gaussian prior on the observations yields $\mathbf{f}^* | X^*, X, \mathbf{f} \, \sim \, \mathcal{N} (\bm{\mu}, \bm{\Sigma})$ where
\vspace{-0.1cm}
{\small
\begin{align}
	\bm{\mu} &= \mathbf{m} - K(X^*,X) K(X,X)^{-1}(\mathbf{f} - \mathbf{m}), \label{mu}\\
	\bm{\Sigma} &=  K(X^*,X^*) - K(X^*,X) K(X,X)^{-1} K(X,X^*). \notag \vspace{-0.1cm}
\end{align}}
\vspace{-0.1cm}
\indent The choice of the kernel $k(\mathbf{x},\mathbf{x}')$ impacts predictions. 
We investigate 4 different kernels:
\vspace{0.1cm}

\begin{itemize}
\item Linear: $k(\mathbf{x}, \mathbf{x}') = \mathbf{x}^T \mathbf{x}'$
\item Cubic: $k(\mathbf{x}, \mathbf{x}') = 3  \left (\left (\mathbf{x}^T \mathbf{x}'  \right )^2 + 2\left ( \mathbf{x}^T \mathbf{x}' \right )^3  \right )$
\item Absolute exponential: $k(\mathbf{x}, \mathbf{x}') = e^{|\mathbf{x}-\mathbf{x}'|} $
\item Squared exponential: $k(\mathbf{x}, \mathbf{x}') = e^{-0.5|\mathbf{x}-\mathbf{x}'|^2} $
\end{itemize}

\vspace{0.1cm}
To initialize the GP search, one needs to compute the F1-score for a certain number of randomly chosen hyperparameter combinations $r$: we investigate what the optimal number is.
We then iterate over the following two steps until a specified maximum number of iterations $t$ is reached. 
First, we find the hyperparameter combination in the test set with the highest F1-score predicted by the GP. Second, we compute the actual F1-score, and move it to the training set.
This process is outlined in Algorithm \ref{algo1}. 
\begin{algorithm} %
\caption{GP search algorithm}\label{algo1}
\begin{algorithmic}
\Function{GP-Regression}{$X^*, X, \mathbf{f}$}

	\State compute $\bm{\mu}$
	according to \eqref{mu}
	\State \Return $\bm{\mu}$
\EndFunction
\vspace{0.1cm}
\Function{GP-Search}{$\mathcal{X} = \{\mathbf{x}_1,\dotsc,\mathbf{x}_s\}, f(\cdot), r, t$}
\State $X \leftarrow (\emptyset)$
\State $X^* \leftarrow (\mathbf{x}_1, \dotsc, \mathbf{x}_s)$
\For{$i = 1, \dotsc, r$}
\State randomly choose $\mathbf{x} \in X^*$  
\State remove $\mathbf{x}$ from $X^*$ 
\State add $\mathbf{x}$ to $X$ and $f(\mathbf{x})$ to $\mathbf{f}$
\EndFor
\For{$i = r+1, \dotsc, t$}
\State $\bm{\mu} \leftarrow$ \Call{GP-Regression}{$X^*, X, \mathbf{f}$}
\State $\hat{j} \leftarrow \displaystyle \argmax_{j = 1, \dotsc, |\bm{\mu}|} \mu_j, \; \mathbf{x} \leftarrow X[j^*]$
\State remove $ \mathbf{x} $ from $X^*$ 
\State add $\mathbf{x}$ to $X$ and $f(\mathbf{x})$ to $\mathbf{f}$ 
\EndFor
\State \Return $\argmax_{\mathbf{x} \in X} f(\mathbf{x})$
\EndFunction
\end{algorithmic}
\end{algorithm}

\input{graphics/kernels.tex}

%% file: graphics/kernels.tex
\begin{figure*}[ht]
\centering
  \subfloat[\vspace{-0.0cm}][DSTC~4]{\includegraphics[height=4.3cm]{{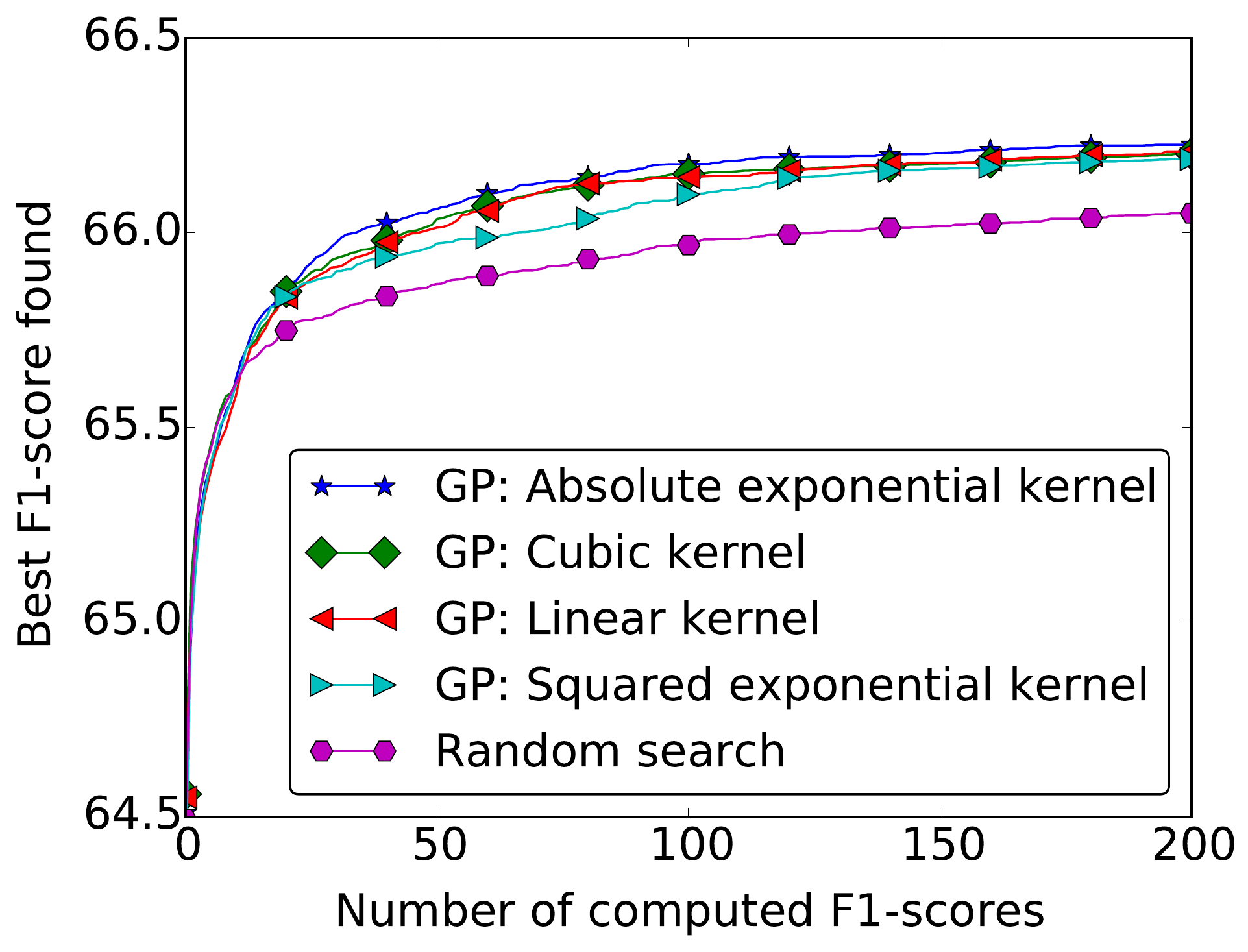}}\label{fig:sub1}} \hspace{0.4cm}
    \subfloat[\vspace{-0.0cm}][MRDA]{\includegraphics[height=4.3cm]{{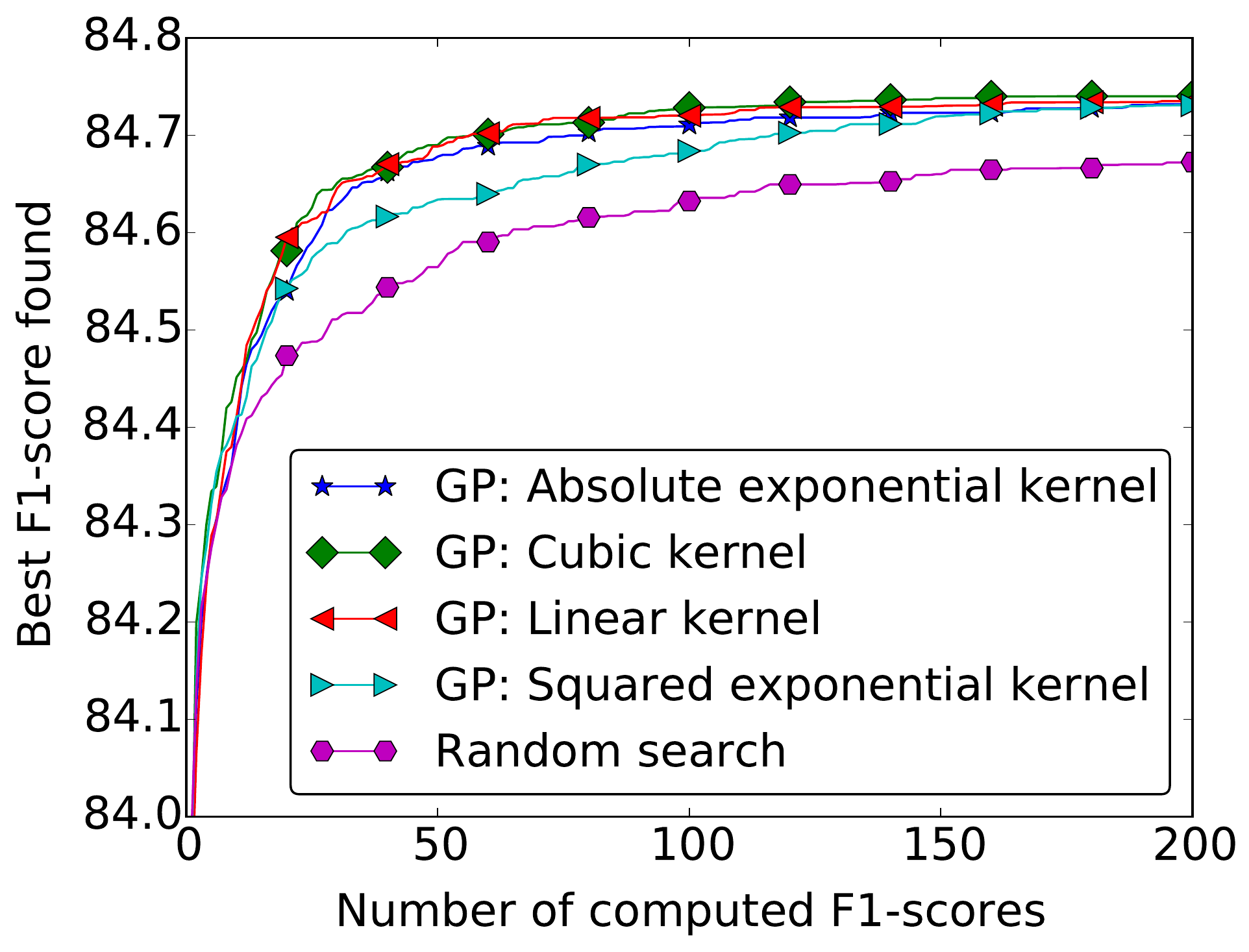}} \label{fig:sub2}} \hspace{0.4cm}%
  \subfloat[\vspace{-0.0cm}][SwDA]{\includegraphics[height=4.3cm]{{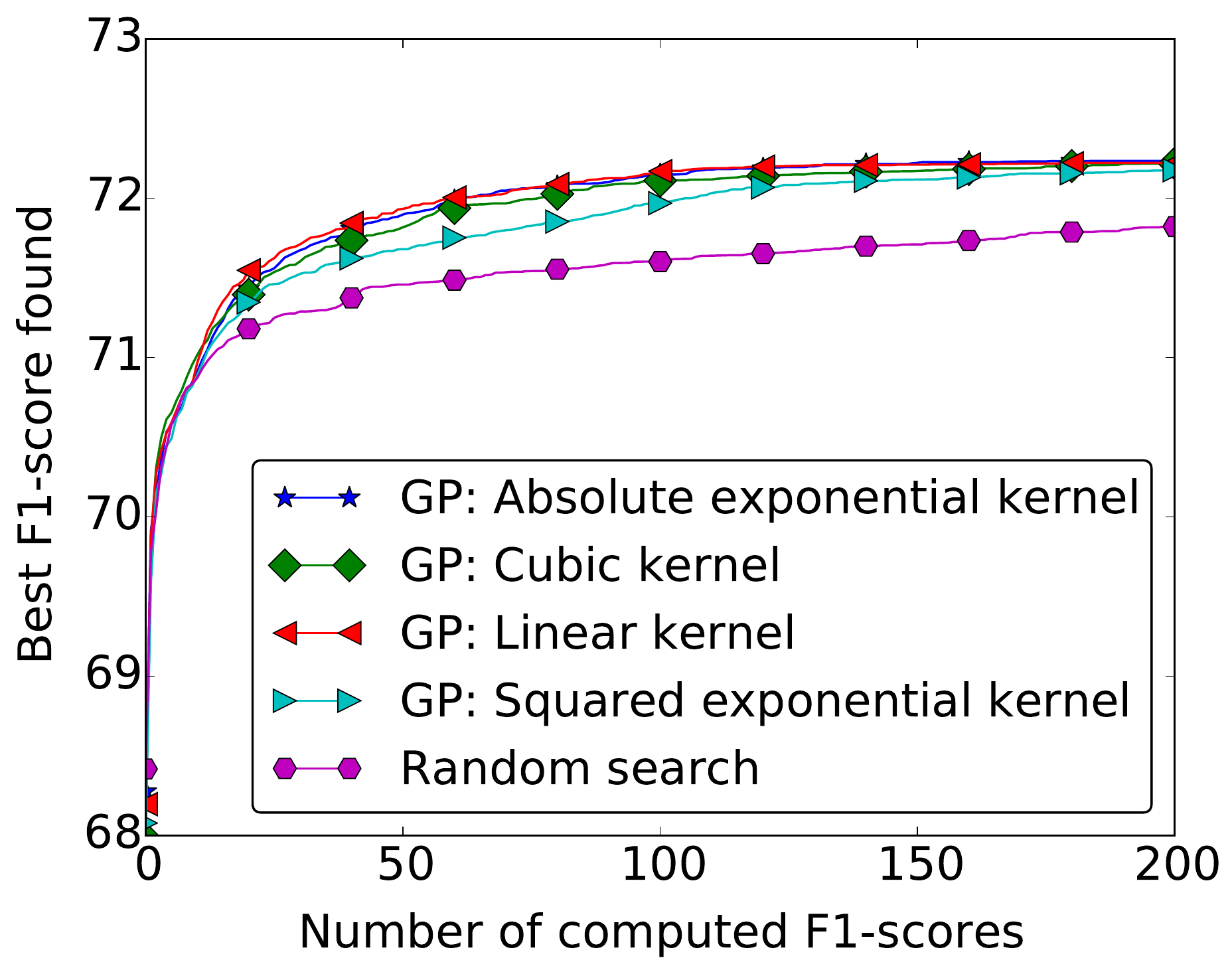}}\label{fig:sub3}}
  \vspace{-0.0cm}
  \caption{Performance of GP search with different kernels and random search for hyperparameter optimization on DSTC~4, MRDA, and SwDA. The x-axis represents the number of hyperparameter combinations for which the F1-score has been computed, and the y-axis shows the best F1-score that has been achieved by at least one of these hyperparameter combinations. 
  Each data point is averaged over 100 runs of the specified search strategy.} \label{fig:kernels}
  \vspace{0.2cm}
\end{figure*}

%% file: experiments.tex
\section{Experiments}
\vspace{-0.1cm}
\subsection{Datasets}

We evaluate the random and GP searches on the dialog act classification task using the Dialog State Tracking Challenge~4 (DSTC~4)~\protect\cite{DSTC4handbook,DSTC4}, ICSI Meeting Recorder Dialog Act (MRDA)~\protect\cite{janin2003icsi,shriberg2004icsi}, and Switchboard Dialog Act (SwDA)~\protect\cite{jurafsky1997switchboard} datasets. 
DSTC~4, MRDA, and SwDA respectively contain 32k, 109k, and 221k utterances, which are labeled with 89, 5, and 43 different dialog acts (we used the 5 coarse-grained dialog acts introduced in~\protect\cite{ang2005automatic} for MRDA).
The train/test splits are provided along with the datasets, and the validation set was chosen randomly except for MRDA, which specifies a validation set.\footnote{See \url{https://github.com/Franck-Dernoncourt/slt2016} for the train, validation, and test splits.}

\vspace{0.0cm}
\subsection{Training} \label{sec:training}
For a given hyperparameter combination, the ANN is trained to minimize the negative log-likelihood of assigning the correct dialog acts to the utterances in the training set, using stochastic gradient descent with the Adadelta update rule~\cite{zeiler2012adadelta}.
At each gradient descent step, weight matrices, bias vectors, and word vectors are updated.
For regularization, dropout is applied after the pooling layer, and early stopping is used on the validation set with a patience of 10 epochs. 
We initialize the word vectors with the 300-dimensional word vectors pretrained with word2vec on Google News
~\cite{mikolov2013efficient,mikolov2013distributed} for DSTC~4, and the 200-dimensional word vectors pretrained with GloVe on Twitter
~\cite{pennington2014glove} for SwDA.

\vspace{-0.2cm}
\subsection{Hyperparameters} \label{sec:search}

For each hyperparameter combination, the reported F1-score is averaged over 5 runs. Table~\ref{tab:hyperparameters-ranges} presents the hyperparameter search space.

\input{tables/hyperparameters3}

\vspace{-0.4cm}

\input{graphics/random-points.tex}

%% file: tables/hyperparameters3.tex
{\renewcommand{\arraystretch}{0.7}

\begin{table}[H]
\small
\begin{center}
\begin{tabular}{>{\centering}m{2.85cm}>{\centering}m{4cm}}
 \toprule
 Hyperparameter & \vspace{0.00cm} Values\tabularnewline
\midrule
 \vspace{0.05cm} Filter size $h$ \vspace{0.05cm} & 3, 4, 5 \tabularnewline
 \vspace{0.05cm} Number of filters $n$ \vspace{0.05cm} & 50, 100, 250, 500, 1000\tabularnewline
 \vspace{0.05cm} Dropout rate $p$ \vspace{0.05cm} & 0.1, 0.2, $\dotsc$ , 0.9\tabularnewline
  \vspace{0.05cm} History size $d_1$ \vspace{0.05cm} & 1, 2, 3\tabularnewline
  \vspace{0.05cm} History size $d_2$ \vspace{0.05cm} & 1, 2, 3\tabularnewline
 \bottomrule
\end{tabular}
\caption{Candidate values for each hyperparameter. Since $h$, $n$, $p$, $d_1$, and $d_2$ can take 3, 5, 9, 3, and 3 different values respectively, there are 1215 ($ = 3 \times 5 \times 9 \times 3 \times 3$) possible hyperparameter combinations.}
\label{tab:hyperparameters-ranges}
\end{center}
\end{table}
}

%% file: graphics/random-points.tex
\begin{figure*}[ht]
\centering
  \subfloat[\vspace{-0.0cm}][DSTC~4]{\includegraphics[height=4.3cm]{{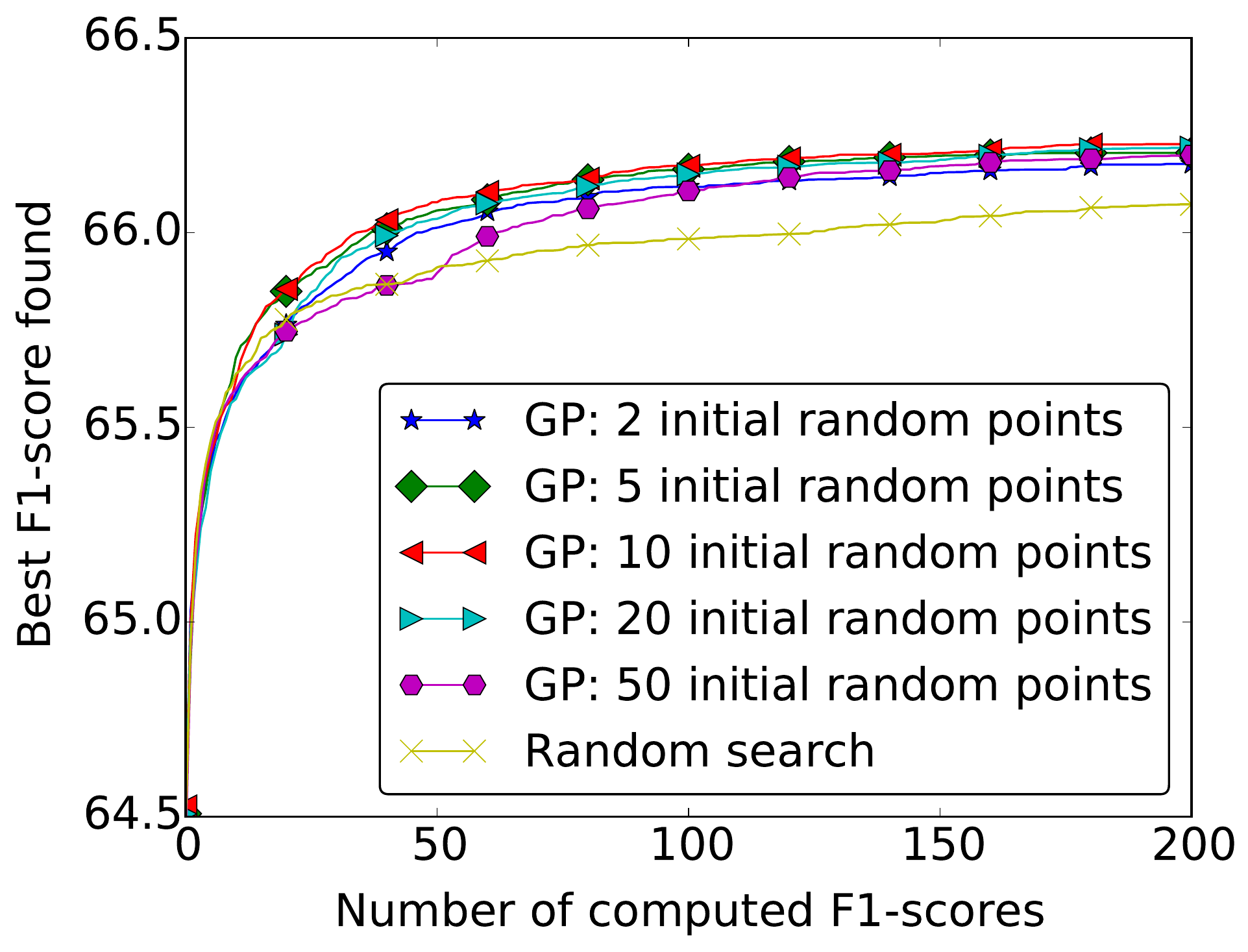}}\label{fig:sub1}} \hspace{0.4cm}
    \subfloat[\vspace{-0.0cm}][MRDA]{\includegraphics[height=4.3cm]{{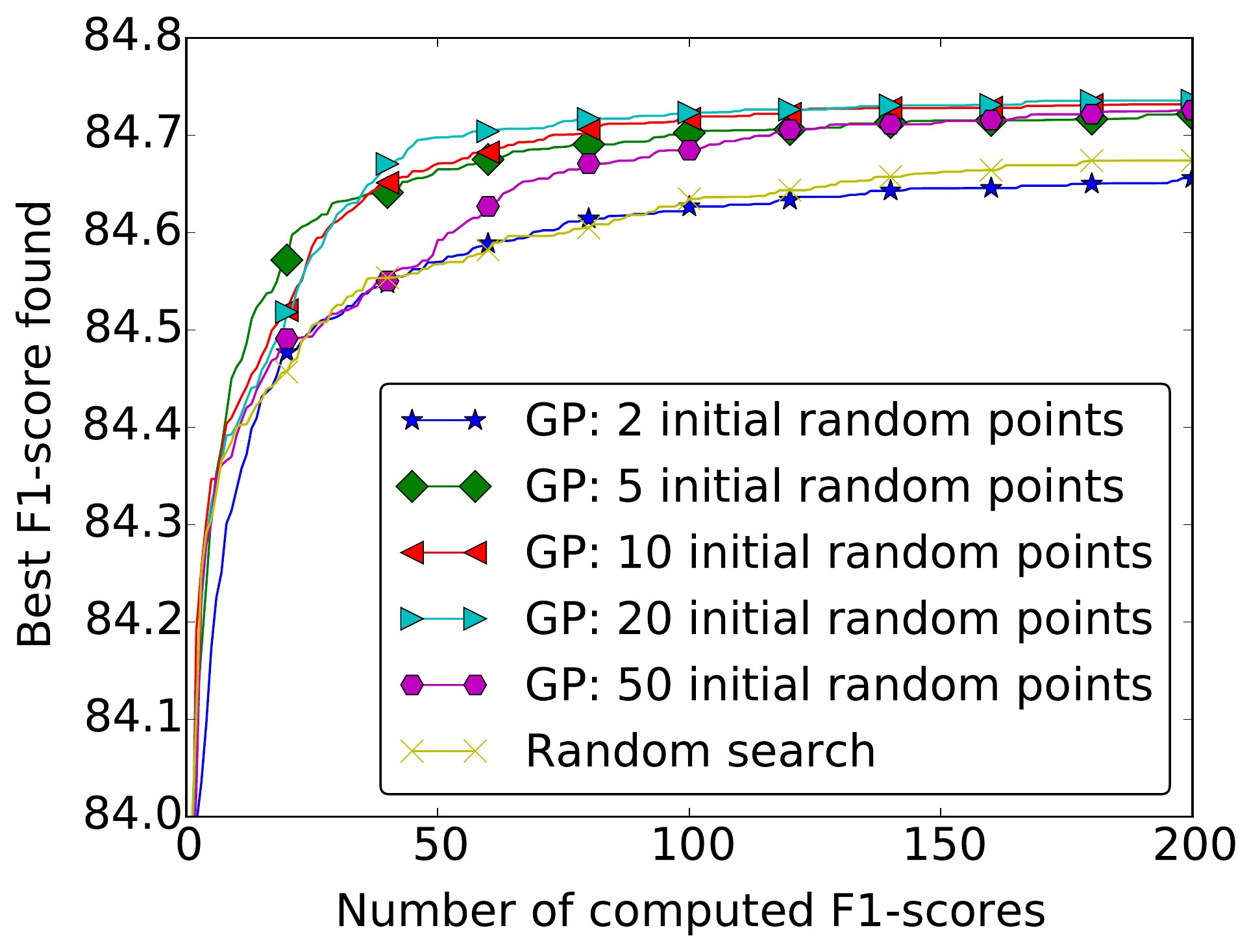}} \label{fig:sub2}} \hspace{0.4cm}%
  \subfloat[\vspace{-0.0cm}][SwDA]{\includegraphics[height=4.3cm]{{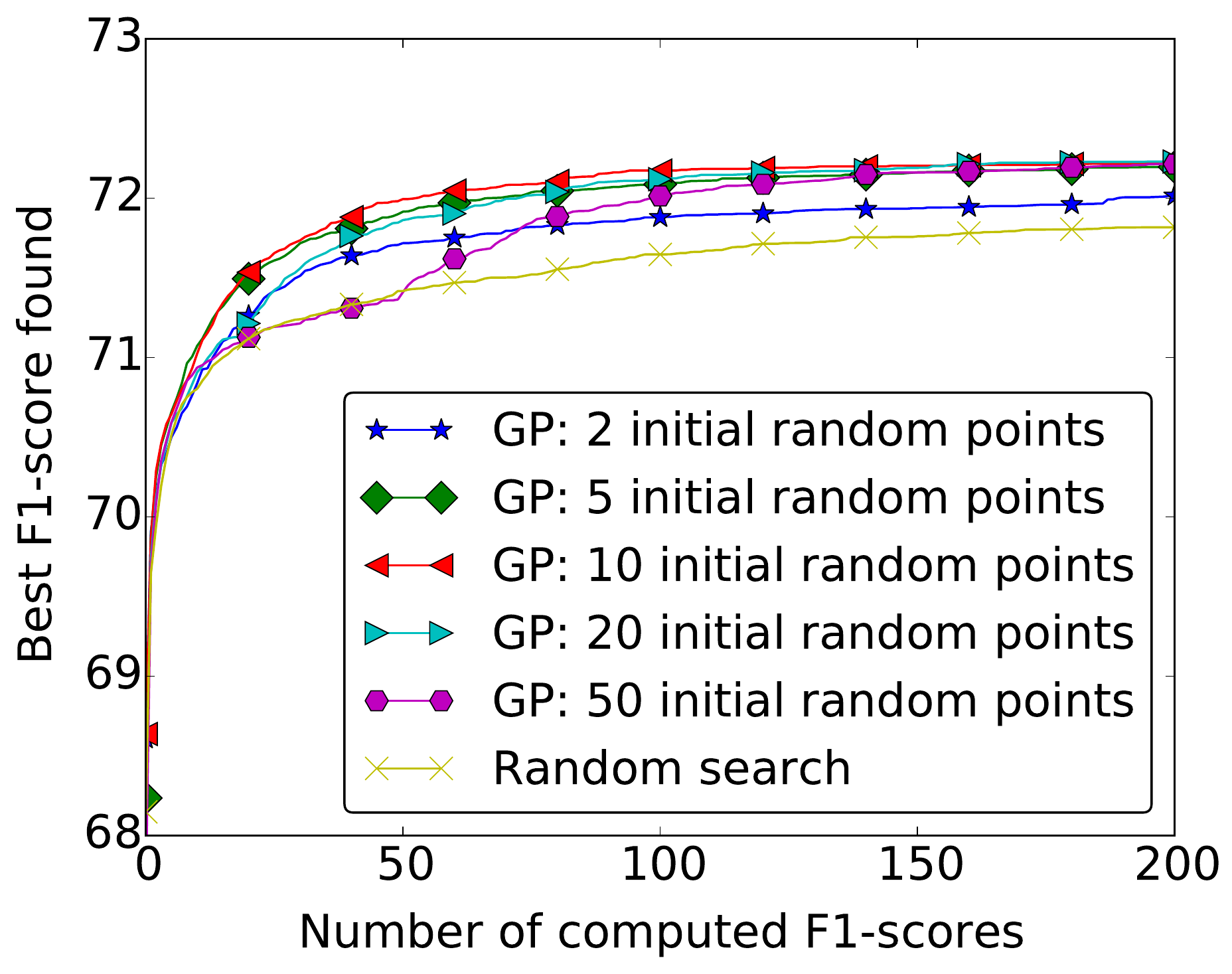}}\label{fig:sub3}}
  \vspace{-0.0cm}
  \caption{Impact of the number of initial random hyperparameter combinations on the GP search. The x-axis represents the number of hyperparameter combinations for which the F1-score has been computed, and the y-axis shows the best F1-score that has been achieved by at least one of these hyperparameter combinations. 
  Each data point is averaged over 100 runs of the specified search strategy.} \label{fig:random}
\end{figure*}

%% file: results.tex
\input{graphics/histogram-tops.tex}

\vspace{-0.5cm}
\section{Results}
\textbf{GP search finds near-optimal hyperparameters faster than random search.}
Figure~\ref{fig:kernels} compares  
the GP searches with different kernels against the random search, which is a natural baseline for hyperparameter optimization algorithms \cite{bergstra2012random}.
On all datasets, the F1-score evaluated using the hyperparameters found by the GP search converges to near-optimal values significantly faster than the random search, regardless of the kernels used.
For example, on SwDA, after computing the F1-scores for 100 different hyperparameter combinations, the GP search reaches on average 72.1, whereas the random search only obtains 71.4. The random search requires computing over 400 F1-scores to reach 72.1: the GP search therefore reduces the computational time by a factor of 4.
This is a significant improvement considering that computing the average F1-scores over 5 runs for 300 extra hyperparameter combinations takes 60 days on a GeForce GTX Titan X GPU.

\textbf{Squared exponential kernel converges more slowly than others.} Even though the GP search with any kernel choice is faster than the random search, some kernels result in better performance than others. The best kernel choice depends on the choice of the dataset, but the squared exponential kernel (a.k.a. radial basis function kernel) consistently converges more slowly, as illustrated by Figure \ref{fig:kernels}. Across the datasets, there was no consistent differences among the linear, absolute exponential, and cubic kernels.

\textbf{The number of initial random points impacts the performances.} As mentioned in Section~\ref{sec:gp}, the GP search starts with computing the F1-score for a certain number of randomly chosen hyperparameter combinations. Figure~\ref{fig:random} shows the impact of this number on all three datasets. The optimal number seems to be around 10 on average, i.e. 1\% of the hyperparameter search space.
When the number is very low (e.g., 2), the GP might fail to find the optimal hyperparameter combinations: it performs significantly worse on MRDA and SwDA.
Conversely, when the number is very high (e.g., 50) it unnecessarily delays the convergence. 

\textbf{GP search often finds near-optimal hyperparameters quickly.} After evaluating the F1-scores with 50 hyperparameter combinations, the GP search finds one of the 5 best hyperparameter combinations almost 80\% of the time on SwDA, as shown in Figure \ref{fig:histogram}, and even more frequently on DSTC~4 and MRDA. After computing 100 hyperparameter combinations, the GP search finds the best one over 70\% of the time, while the random search stumbles upon it less 10\% of the time.

\begin{figure*}[!ht]
  \centering
      \includegraphics[width=\textwidth]{{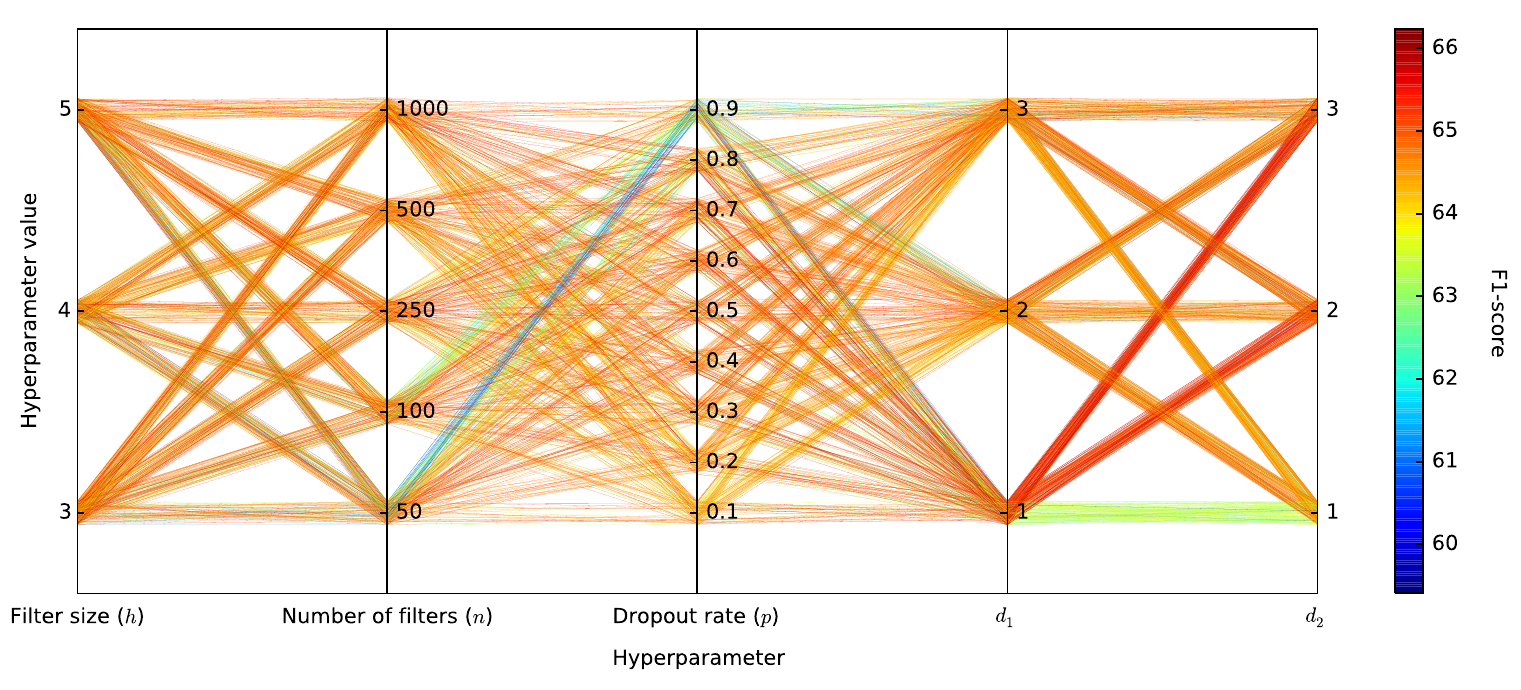}}

  \caption{Parallel coordinate plot of all 1215 hyperparameter combinations for DSTC~4. Each hyperparameter combination in 5-dimensional search space is shown as a polyline with vertices on the parallel axes, each of which represents one of the 5 hyperparameter. The position of the vertex on each axis indicates the value of the corresponding hyperparameter. The color of each polyline reflects the F1-score obtained using the hyperparameter combination corresponding to the polyline.
  \vspace{-0.0cm}} 
  \label{fig:parallel_coordinates}
\vspace{-0.0cm}
\end{figure*}

\vspace{-0.0cm}
\begin{figure}[!ht]
  \centering
      \includegraphics[width=0.5\textwidth]{{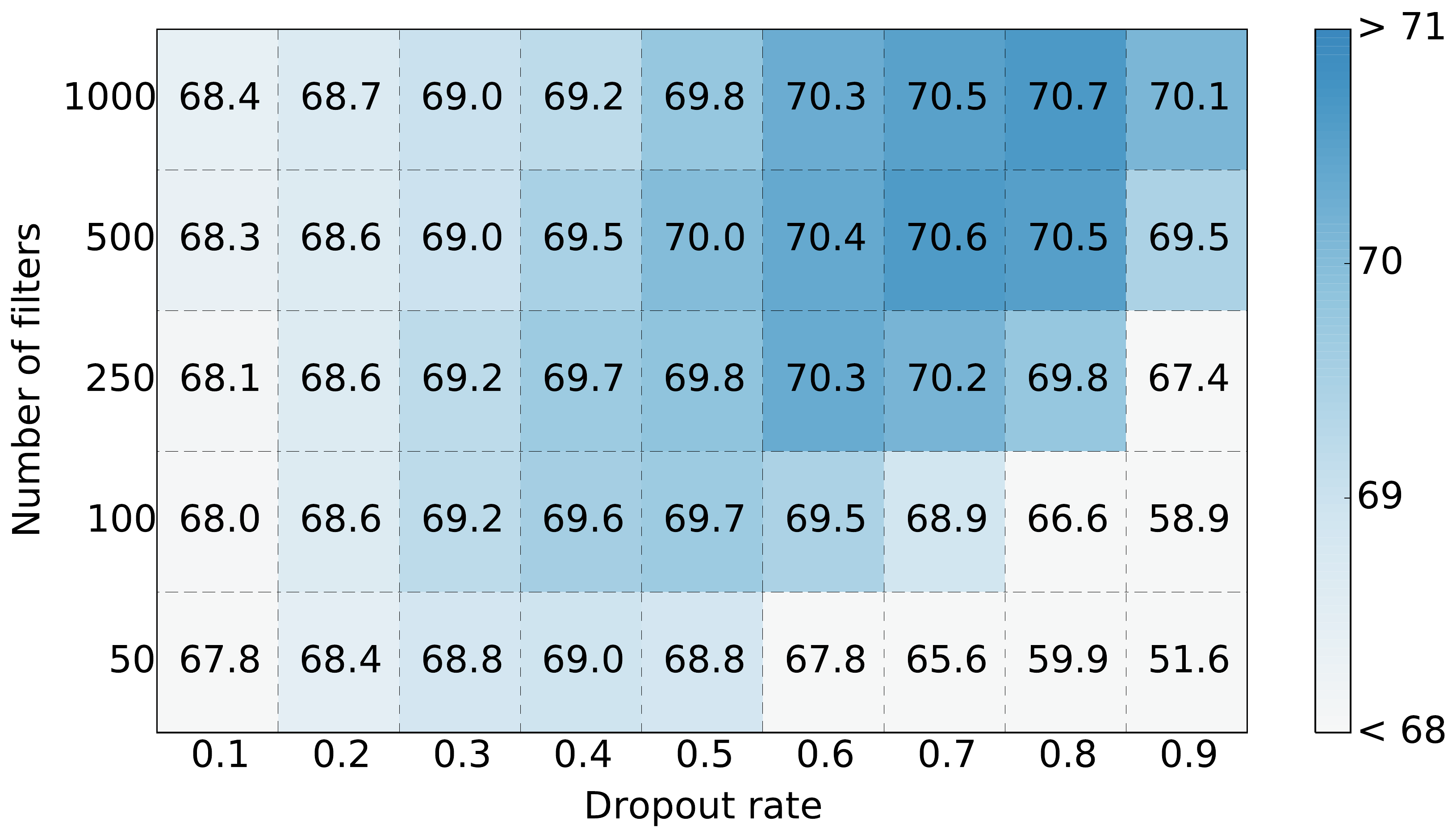}}

  \caption{Heatmap of the F1-scores on SwDA as the number of filters and the dropout rate vary. F1-scores are averaged over all possible values of the other hyperparameters: as a result, F1-scores can be lower than the ones in Figure~\ref{fig:kernels}. \vspace{-0.0cm}} 
  \label{fig:heatmap}
\end{figure}

~\newline
~\newline
~\newline
\textbf{Simple heuristics may not find optimal hyperparameters well.}
Compared to the previous state-of-the-art results that use the same model optimized manually~\cite{lee2016sequential}, the GP search found more optimal hyperparameters, improving the F1-score by 0.5 ($=66.3-65.8$), 0.1 ($=84.7-84.6$), and 0.7 ($=72.1-71.4$) on DTSC~4, MRDA, and SwDA, respectively. In \cite{lee2016sequential}, the hyperparameters were optimized by varying one hyperparameter at a time while keeping the hyperparameters fixed.
Figures~\ref{fig:parallel_coordinates} and \ref{fig:heatmap} demonstrate that optimizing each hyperparameter independently
might result in a suboptimal choice of hyperparameters. 
Figure~\ref{fig:parallel_coordinates} illustrates that the optimal choice of hyperparameters is impacted by the choice of other hyperparameters. For example, a higher number of filters works better with a smaller dropout probability, and conversely a lower number of filters yields better results when used with a larger dropout probability. 
Figure~\ref{fig:heatmap} shows that, for instance, if one had first fixed the number of filters to be 100 and optimized the dropout rate, one would have found that the optimal dropout rate is 0.5. Then, fixing the dropout rate at 0.5, one would have determined that 500 is the optimal number of filters, thereby obtaining an F1-score of 70.0, which is far from the best F1-score (70.7).
\newpage
The faster convergence of the GP search may stem from the capacity of the GP to leverage the patterns in the F1-score landscape such as the one shown in Figure~\ref{fig:heatmap}. 
The random search cannot make use of this regularity.

%% file: graphics/histogram-tops.tex
\begin{figure*}
\centering
  \subfloat[\vspace{-0.0cm}][]{\includegraphics[height=4.3cm]{{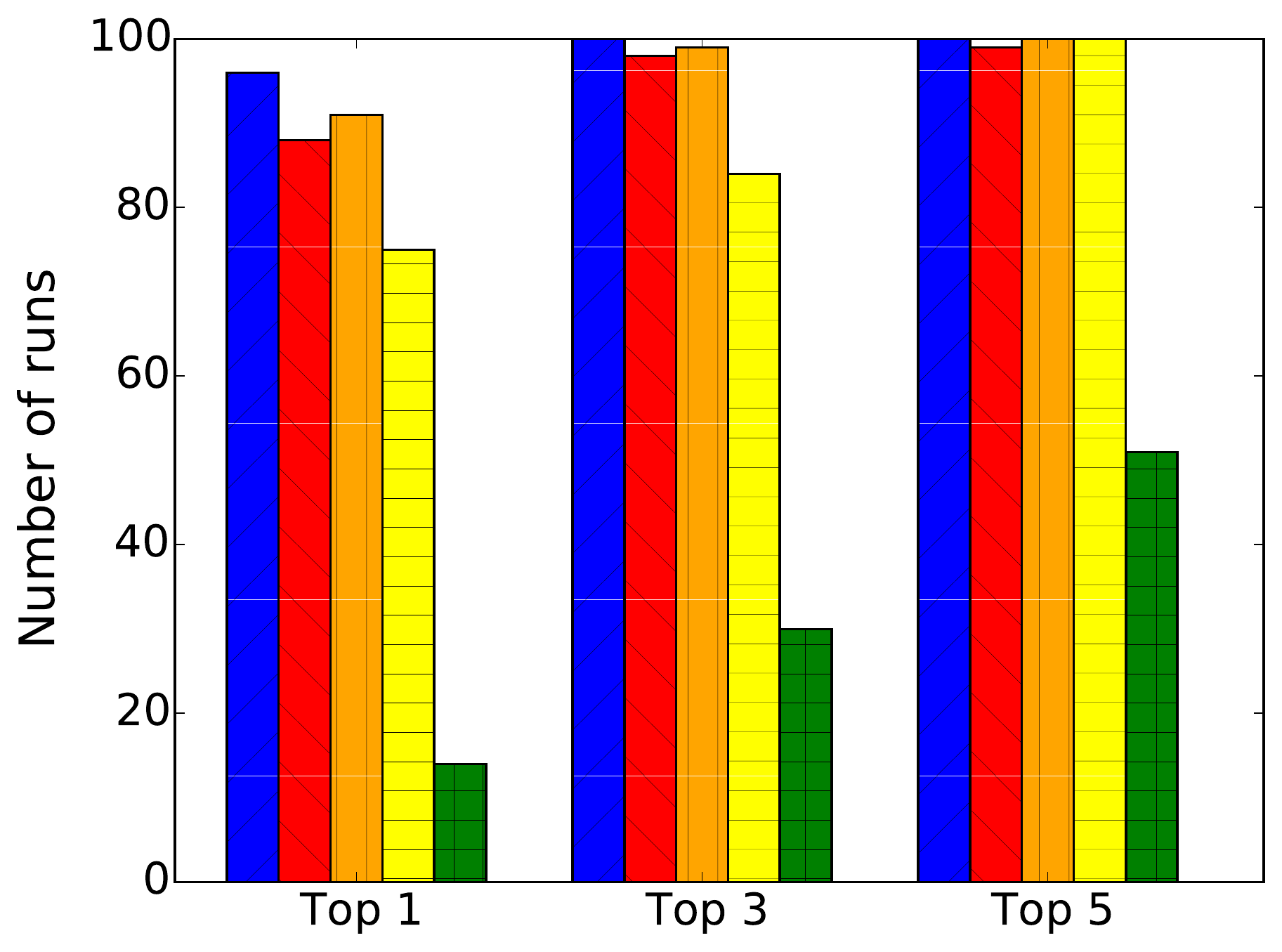}}\label{fig:sub1}} \hspace{0.6cm}
    \subfloat[\vspace{-0.0cm}][]{\includegraphics[height=4.3cm]{{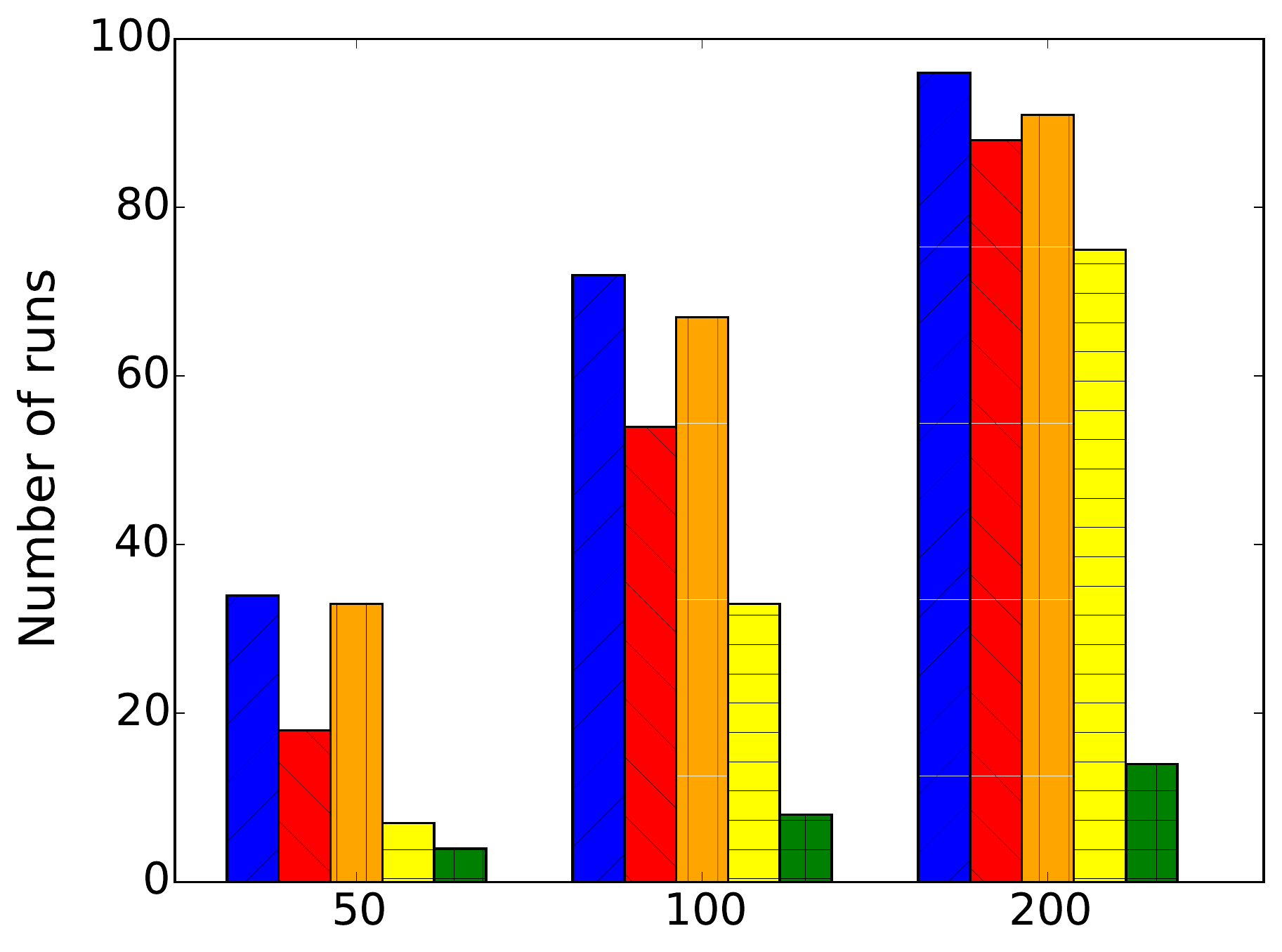}} \label{fig:sub2}} \hspace{0.4cm}%
 \subfloat{\includegraphics[height=4.3cm]{{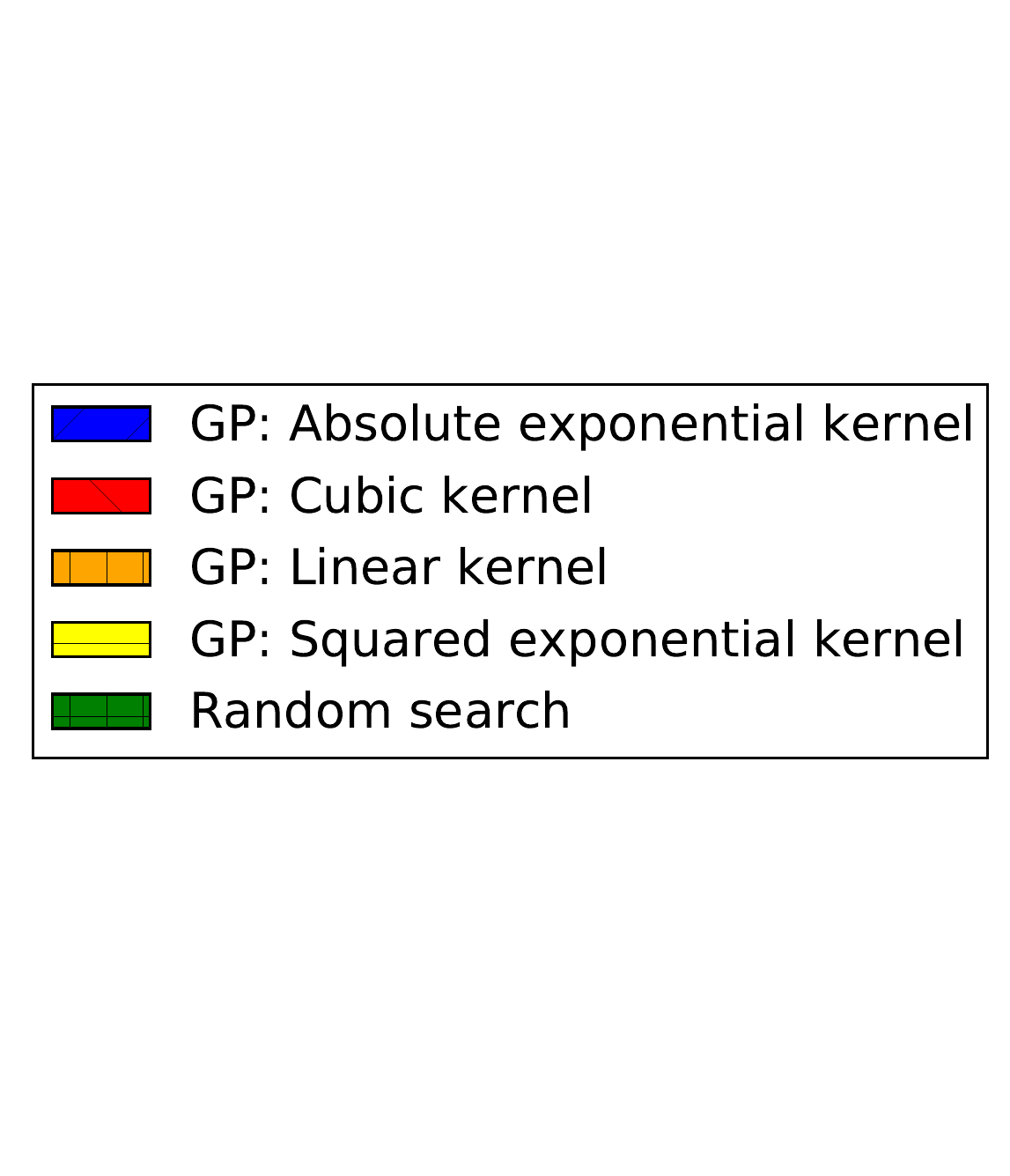}}\label{fig:sub3}}

  \vspace{-0.0cm}
  \caption{Finding near-optimal hyperparameter combinations on SwDA. Figure (a) shows how many times out of 100 runs each search strategy found a hyperparameter combination that is among the top 1, 3, and 5 best performing hyperparameter combinations. Figure (b) shows how many times out of 100 runs each search strategy found the best hyperparameter combination after evaluating 50, 100, and 200 hyperparameter combinations.} \label{fig:histogram}
  \vspace{-0.1cm}
\end{figure*}

%% file: conclusion.tex
\section{Conclusion}

In this paper we addressed the commonly encountered issue of tuning ANN hyperparameters.
Towards this purpose, we explored a strategy based on GP to automatically pinpoint optimal or near-optimal ANN hyperparameters. 
We showed that the GP search requires 4 times less computational time than random search on three datasets, and improves the state-of-the-art results by efficiently finding the optimal hyperparameter combinations.
While the choices of the kernels and the number of initial random points impact the performance of the GP search, our findings show that it is more efficient than the random search regardless of these choices.  
The GP search can be used for any ordinal hyperparameter; it is therefore a useful technique when developing ANN models for NLP tasks.